\documentclass[journal]{IEEEtran}

\usepackage{amsmath,amsfonts}
\usepackage{algorithmicx}
\usepackage{algorithm}
\usepackage{array}
\usepackage[caption=false,font=normalsize,labelfont=sf,textfont=sf]{subfig}
\usepackage{textcomp}
\usepackage{stfloats}
\usepackage{url}
\usepackage{verbatim}
\usepackage{graphicx}
\usepackage{cite}

\usepackage{graphicx}
\usepackage{tabularx}
\usepackage{booktabs}
\usepackage{diagbox}
\usepackage{amsfonts}
\usepackage{pifont}
\usepackage{algpseudocode}
\usepackage[dvipsnames]{xcolor}
\usepackage{multirow} 
\usepackage{makecell} 
\usepackage{array,adjustbox}
\usepackage{cleveref}
\usepackage{tabularx}
\usepackage{anyfontsize}

\newcommand{\figlabel}{Fig.}

\newcommand{\fig}[1]{\figlabel{} \ref{#1}}

\newcommand{\tablabel}{Table}

\newcommand{\tab}[1]{\tablabel{} \ref{#1}}

\newcommand{\sectlabel}{Section}
\newcommand{\sect}[1]{\sectlabel{} \ref{#1}}

\newcommand{\sheh}{Shoaib Ehsan}
\newcommand{\shehEMe}{sehsan@essex.ac.uk}
\newcommand{\shehEMs}{s.ehsan@soton.ac.uk}

\newcommand{\mm}{Michael Milford}
\newcommand{\mmEM}{michael.milford@qut.edu.au}

\newcommand{\timur}{Timur Ismagilov}
\newcommand{\timurEM}{ti1e24@soton.ac.uk}

\newcommand{\sdr}{Sarvapali D. Ramchurn}
\newcommand{\sdrEM}{sdr1@soton.ac.uk}

\newcommand{\tvtn}{Tan Viet Tuyen Nguyen}
\newcommand{\tvtnEM}{tuyen.nguyen@soton.ac.uk}

\newcommand{\sm}{Shakaiba Majeed}
\newcommand{\smEM}{shakaiba33@hanyang.ac.kr}

\title{\LARGE 
One Channel to Rule Them All: Rethinking Input Representation for Visual Place Recognition
}

\author{\timur{}$^{1}$*, \sm{}$^{2}$, \mm{}$^{3}$, \tvtn{}$^{1}$ \\
\sdr{}$^{1}$, \sheh{}$^{1,4}$ 

\thanks{*Corresponding author.}%
\thanks{$^{1}$\timur{}, \tvtn{}, \sdr{} and \sheh{} are with the School of Electronics and Computer Science, University of Southampton, SO17 1BJ Southampton, U.K. {\tt\footnotesize \timurEM{}, \tvtnEM{}, \sdrEM{}, \shehEMs{}}}%
\thanks{$^{2}$\sm{} is with the  Department of Computer Science and Engineering, Hanyang University, Seoul, South Korea. {\tt\footnotesize \smEM{}}}%
\thanks{$^{3}$\mm{} is with the QUT Centre for Robotics, School of Electrical
Engineering and Robotics, Brisbane, QLD 4000, Australia. {\tt\footnotesize \mmEM{}}}%
\thanks{$^{4}$\sheh{} is also with the School of Computer Science and Electronic Engineering, University of Essex, Colchester, CO4 3SQ, U.K.  {\tt\footnotesize \shehEMe{}}}%
}

\begin{document}

\maketitle
\thispagestyle{empty}
\pagestyle{empty}

\begin{abstract}

Visual Place Recognition (VPR) is fundamental to long-term robot localization and SLAM, yet current systems overwhelmingly rely on RGB input, implicitly assuming color is necessary for global place recognition. We challenge this assumption, investigating the role of chromatic information across training regimes, model architectures and standard benchmarks under real-world appearance variation. We find that grayscale matches RGB performance generally and outperforms it under severe appearance shifts where color invariance is insufficiently learned, while color provides meaningful gains only where persistent and discriminative chromatic cues are present. Across selected benchmarks, a fully gray-trained MixVPR model achieves an average 82.4\% Recall@1 compared to 81.2\% for its RGB counterpart. In some cases, lightweight grayscale variants with 60\% fewer parameters can outperform heavier RGB models. Grayscale further offers practical advantages in storage, bandwidth and alignment with resource-constrained systems. We conclude that for global VPR where scenes vary across illumination, weather, season and setting, color contributes minimally, and grayscale alone is sufficient for reliable place recognition.

\end{abstract}


\section{INTRODUCTION}
\label{sec:intro}
\IEEEPARstart{V}{isual} Place Recognition (VPR) is a core component of mobile robotics, enabling navigation and loop-closure detection for Simultaneous Localization And Mapping (SLAM). Unlike object-centric tasks that focus on local discriminative cues for detection or recognition, VPR operates on global scene representations, encoding broad spatial context into compact descriptors. State-of-the-Art (SotA) VPR methods predominantly rely on color imagery, inheriting the conventions of foundational models like ResNet and DINOv2 which provide robust low-level features transferred from large-scale training on ImageNet and LVD-142M datasets. The assumption that color input is required for VPR has become ubiquitous despite limited evaluation of its contribution to performance or its practical implications within VPR systems. 

Recognizing the same place under severe perceptual aliasing is a fundamental challenge in VPR. Appearance changes, including seasonal shifts, weather, illumination, camera viewpoint, dynamic objects and long-term structural changes, require robust models capable of broad generalization. Among these, color variation introduces a degree of freedom that, without sufficient representation during training, can degrade model performance. In contrast, grayscale imagery is centered on luminance-based structure, which may offer more robust place representations under such conditions, while also reducing the input dimensionality, storage requirements for data, and memory utilization across training and inference pipelines. Despite this, the reliance on RGB input in modern VPR architectures remains largely unexamined. Current VPR approaches operate on RGB input driven by the design of their pretrained backbones \cite{eigenplaces, cosplace, netvlad, mixvpr, cricavpr, megaloc}, yet this raises the question of whether color is fundamentally necessary for localization, and whether such models remain susceptible to color-shift induced degradation. Additionally, VPR datasets are almost exclusively stored, trained and evaluated in RGB format \cite{gsvcities, pitts, sf-landmark}, leaving potential gains in data efficiency, memory footprint and edge deployability largely unexplored. 

\begin{figure}[!t]
    \centering
    \includegraphics[width=\columnwidth{}]{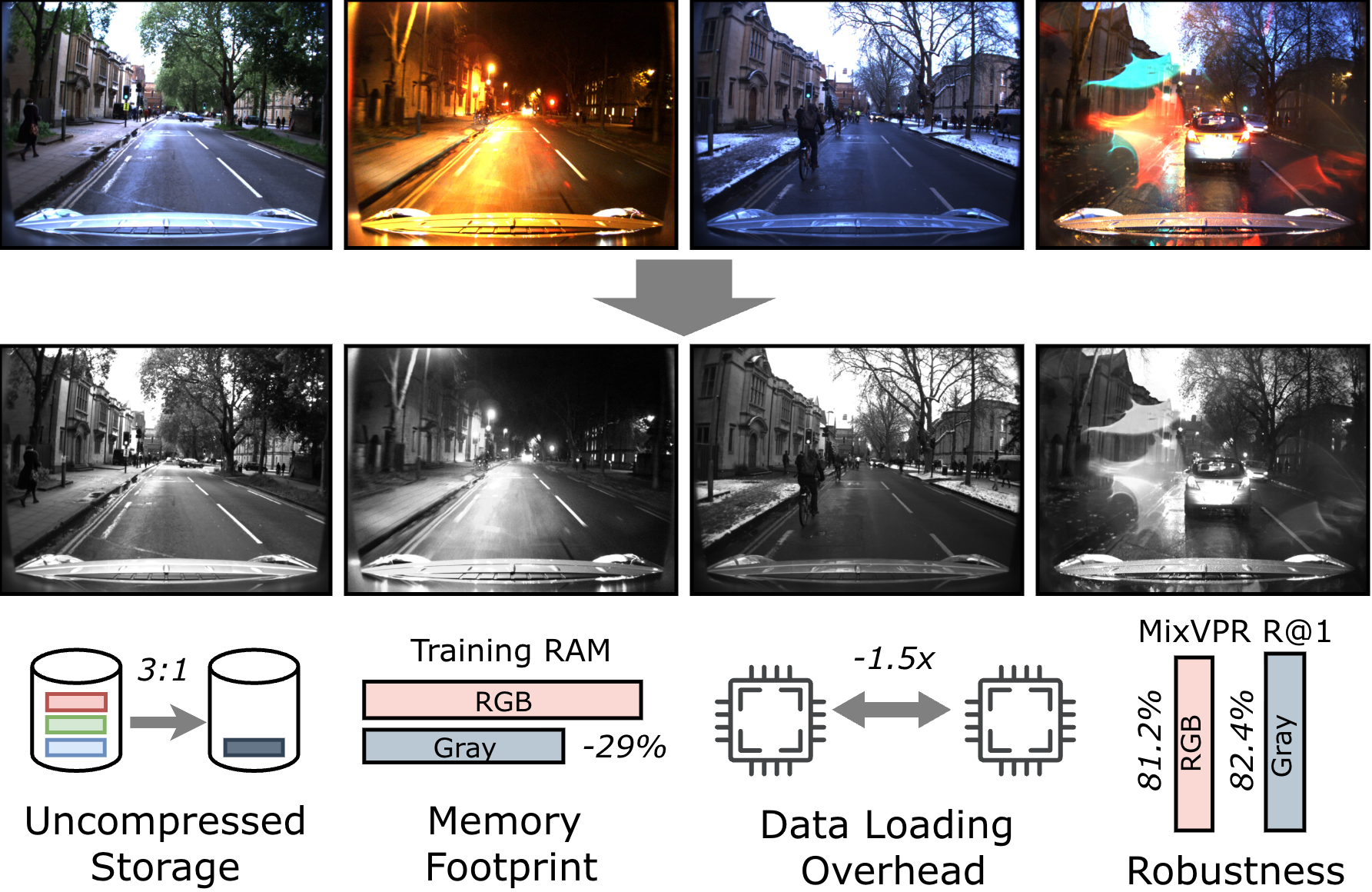}
    \vspace{-2em}
    \caption{We investigate the role of color in VPR, examining its impact on retrieval performance across appearance conditions and the practical considerations of grayscale input for training and deployment.}
    \vspace{-1em}
    \label{fig:Mapping}
\end{figure}

The role of color in both nature and computer vision is widely investigated. Evidence from both suggests its utility is task and data-dependent. In object-centric studies, color imagery has been shown to benefit the classification of faces, edges and object categories such as food \cite{singh2020colours, doi:10.1126/science.adk9587, BRAMAO2011244, Brosseau20042020, 10.1371/journal.pcbi.1007398}. Other works report more nuanced findings, such as indifference in deep face recognition \cite{10887263}, over-sensitivity to hue shifts \cite{doi:10.1126/science.adk9587}, grayscale superiority on texture-dominant data \cite{10.1007/978-3-030-31332-6_27} and broader effects on human perception such as memory retrieval and decision-making under degraded conditions \cite{ 10.1167/8.16.12, DELORME20002187, doi:10.1068/p3376, Ehinger2008}. Within robotic vision, the role of color in VPR has received comparatively little attention. Given the appearance variability inherent to real-world deployment, where color can be a distractor rather than a discriminative signal, we question whether three-channel color input is strictly necessary for VPR with existing models. We make the following contributions: 

\begin{itemize}
    \item An evaluation of RGB and grayscale imagery across SotA models. We find that recent off-the-shelf RGB-trained models show only marginal benefit from color, suggesting that the VPR objective is largely structure and luminance-driven, with color providing marginal support only where consistent chromatic cues are present. 
    
    \item An empirical investigation of the role of color in VPR under different training regimes using MixVPR. When controlling for training, grayscale achieves competitive performance generally and outperforms RGB under challenging appearance shifts, exposing insufficient color invariance in current training conventions.
    
    \item A discussion of practical advantages with grayscale input, including reduced storage, data-loading overhead and memory footprint, alongside its suitability for resource-constrained deployment and embedded sensor modalities.
\end{itemize}

The rest of the paper is organized as follows: \sect{sec:relatedwork} provides an overview of VPR literature and color in recognition tasks. \sect{sec:expsetup} outlines our experimental setup. Results are presented and discussed in \sect{sec:results} and \sect{sec:disc}. Finally, we conclude and discuss future work in \sect{sec:conclusion}.

\begin{figure*}[!ht]
  \centering
  \includegraphics[width=0.85\textwidth]{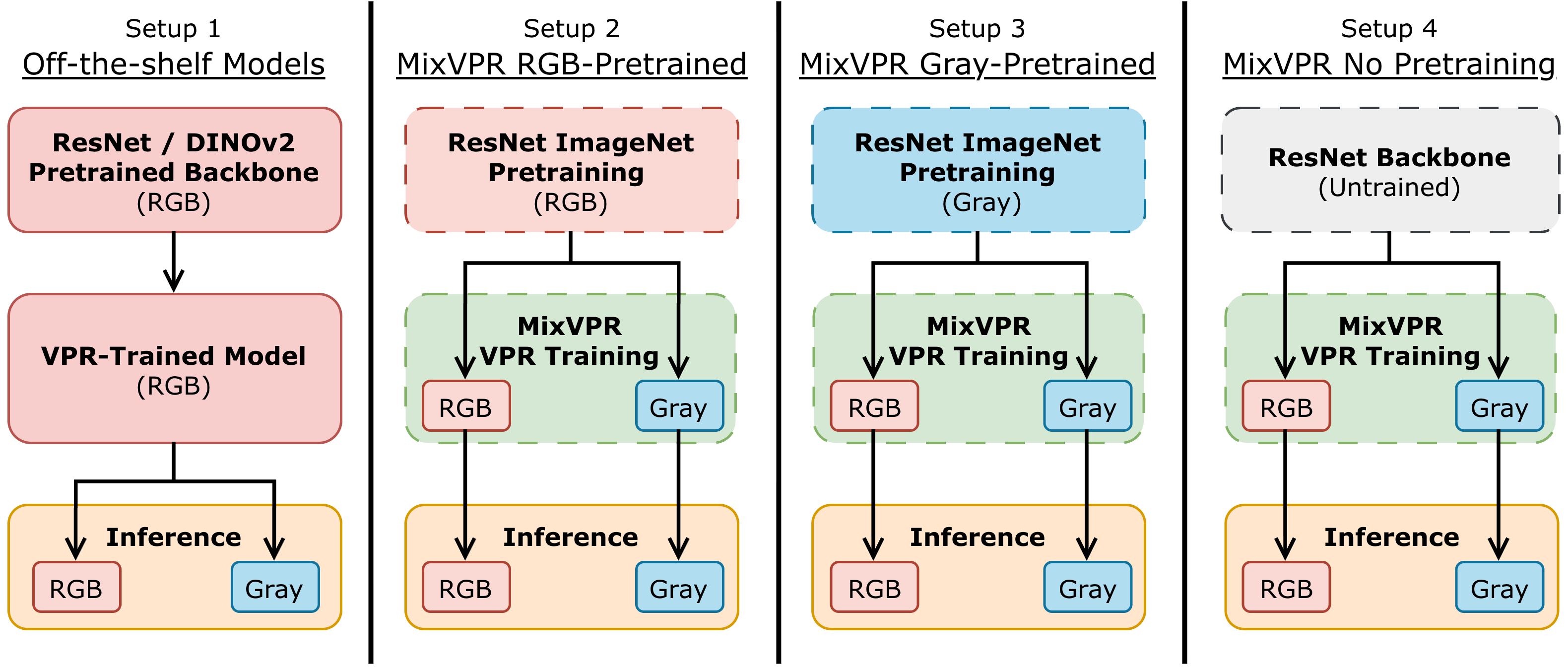}
  \vspace*{-1em}
  \caption{Overview of experimental setups. Setup 1 evaluates off-the-shelf RGB-trained models directly. Setups 2 \& 3 fine-tune MixVPR on GSV-Cities using RGB and grayscale-pretrained ResNet backbones respectively, across both modalities. Setup 4 trains MixVPR from scratch without ImageNet pretraining, also across both modalities. All MixVPR training uses identical hardware, hyperparameters and data splits.}
  \label{fig:blockdiag}
  \vspace*{-1em}
\end{figure*}
\begin{table}[!t]
\centering
\caption{Off-the-shelf VPR models evaluated in this work.}
\label{tab:models}
\resizebox{\columnwidth}{!}{%
\begin{tabular}{@{}lccccc@{}}
    \toprule
    \textbf{Model} & \textbf{Backbone} & \textbf{Pretrain} & \textbf{VPR Train} & \textbf{Params} & \textbf{Dim} \\
    \midrule
    EigenPlaces~\cite{eigenplaces} & ResNet-50 & ImageNet & SF-XL & 27.7M & 2048 \\
    MixVPR~\cite{mixvpr} & ResNet-50 & ImageNet & GSV-Cities & 10.9M & 4096 \\
    CricaVPR~\cite{cricavpr} & ViT-B/14 & LVD-142M & GSV-Cities & 106.8M & 10752 \\
    SALAD~\cite{Izquierdo_CVPR_2024_SALAD} & ViT-B/14 & LVD-142M & GSV-Cities & 88.0M & 8448 \\
    BoQ~\cite{Ali-bey_2024_CVPR} & ViT-B/14 & LVD-142M & GSV-Cities & 95.2M & 12288 \\
    \bottomrule
\end{tabular}}
\vspace{-1em}
\end{table}

\section{RELATED WORK}
\label{sec:relatedwork}
\vspace{-1pt}
\textbf{Visual Place Recognition.} Retrieval-based VPR represents places as compact feature descriptors, using nearest-neighbor retrieval for localization. Early approaches extracted handcrafted local features from grayscale imagery (SIFT \cite{sift}, ORB \cite{orb}), aggregated into global representations via VLAD \cite{vlad}, BoW \cite{bow} or GeM \cite{radenovic2018fine}. Deep learning methods have largely superseded these. MixVPR \cite{mixvpr} uses ResNet50 with channel feature mixing. CosPlace \cite{cosplace} and EigenPlaces \cite{eigenplaces} both use ResNet50 with GeM pooling, differing in their scene partitioning strategy. More recent ViT-based models show stronger retrieval performance: CricaVPR \cite{cricavpr} applies a DINOv2 backbone with cross-image correlation. SALAD \cite{Izquierdo_CVPR_2024_SALAD} reformulates NetVLAD's \cite{netvlad} soft-assignment as optimal transport on DINOv2 features. BoQ \cite{Ali-bey_2024_CVPR} uses learnable global queries with cross-attention, and MegaLoc \cite{megaloc} combines diverse training strategies on a SALAD architecture. Training efficiency is addressed via hard data sampling strategies: CliqueMining \cite{10.1007/978-3-031-73464-9_15} constructs a geographic-descriptor graph to sample hard image pairs, and Global Proxy Mining (GPM) \cite{Ali-Bey_2022_BMVC} uses training proxy descriptors for hard batch construction.

Despite increasing performance over time, the role of color in VPR has received little attention. RGB has been adopted by default with pretrained backbones, yet neither ResNet's object-centric ImageNet supervision nor DINOv2's self-supervised pretraining is guaranteed to capture chromatic variability at VPR inference. Practically, grayscale reduces uncompressed storage by two-thirds relative to RGB, which is significant for terabyte-scale datasets, and proportionally reduces host-to-device transfer, improving training efficiency \cite{10887263}. Edge deployment and TinyML platforms \cite{10284551} operate under tight memory budgets, and even lightweight vSLAM and VPR pipelines have required offline preprocessing to meet real-time constraints on resource-constrained platforms \cite{drones5040121, lightwork}. 

\newcommand{\rot}[1]{\rotatebox[origin=lB]{90}{\hspace{1pt}#1\hspace{4pt}}}
\newcolumntype{Y}{>{\centering\arraybackslash}X}
\newlength{\tablew}
\setlength{\tablew}{\textwidth}
\newlength{\firstcolw}
\setlength{\firstcolw}{1.8cm}
{
\begin{table*}[!ht]
\footnotesize
\centering
\caption{Recall@1 across Nordland season pairs, evaluated on RGB and Gray using \textbf{off-the-shelf Models}.}
\label{tab:nordland_OTS}
\setlength{\tabcolsep}{1.9pt}
\renewcommand{\arraystretch}{1.0}
\begin{tabularx}{\tablew}{|p{\firstcolw}|c|YYYY|YYYY|YYYY|YYYY|YYYY|}
\hline
& \textbf{Model} $\rightarrow$ &
\multicolumn{4}{c|}{\textbf{EigenPlaces}} &
\multicolumn{4}{c|}{\textbf{MixVPR}} &
\multicolumn{4}{c|}{\textbf{SALAD}} &
\multicolumn{4}{c|}{\textbf{CricaVPR}} &
\multicolumn{4}{c|}{\textbf{BoQ}} \\
\cline{2-22}
\makecell[l]{\textbf{Reference} $\downarrow$} &
\makecell[b]{\textbf{Query} $\rightarrow$ \\[0.35em] \textbf{Eval. Mode} $\downarrow$} &
\rot{\textbf{Fall}} & \rot{\textbf{Spring}} & \rot{\textbf{Summer}} & \rot{\textbf{Winter}} &
\rot{\textbf{Fall}} & \rot{\textbf{Spring}} & \rot{\textbf{Summer}} & \rot{\textbf{Winter}} &
\rot{\textbf{Fall}} & \rot{\textbf{Spring}} & \rot{\textbf{Summer}} & \rot{\textbf{Winter}} &
\rot{\textbf{Fall}} & \rot{\textbf{Spring}} & \rot{\textbf{Summer}} & \rot{\textbf{Winter}} &
\rot{\textbf{Fall}} & \rot{\textbf{Spring}} & \rot{\textbf{Summer}} & \rot{\textbf{Winter}} \\
\hline\hline
\multirow{2}{*}{\textbf{Fall}}
& RGB  & --- & \textbf{89.0} & \textbf{93.5} & \textbf{63.1} & --- & \textbf{95.4} & \textbf{95.9} & \textbf{81.7} & --- & \textbf{97.4} & \textbf{97.3} & \textbf{88.9} & --- & \textbf{98.6} & \textbf{98.4} & \textbf{96.1} & --- & \textbf{97.9} & \textbf{97.2} & \textbf{93.0} \\

& Gray & --- & 81.1 & 90.3 & 53.4 & --- & 93.3 & 94.2 & 80.3 & --- & 96.7 & 96.4 & 87.7 & --- & 98.2 & 97.8 & 94.1 & --- & 97.4 & 97.0 & 91.4 \\
\hline

\multirow{2}{*}{\textbf{Spring}}
& RGB  & \textbf{80.5} & --- & \textbf{73.1} & \textbf{68.5} & \textbf{88.1} & --- & \textbf{83.2} & 89.4 & \textbf{95.0} & --- & \textbf{93.0} & 92.2 & \textbf{98.0} & --- & \textbf{95.8} & \textbf{98.3} & 96.2 & --- & \textbf{94.7} & 95.6 \\

& Gray & 70.0 & --- & 64.2 & 59.6 & 84.7 & --- & 80.8 & \textbf{89.8} & 94.5 & --- & 91.4 & \textbf{93.4} & 97.6 & --- & 93.7 & 97.5 & \textbf{96.4} & --- & 93.4 & \textbf{96.2} \\
\hline

\multirow{2}{*}{\textbf{Summer}}
& RGB  & \textbf{94.4} & \textbf{82.7} & --- & \textbf{60.1} & \textbf{95.8} & \textbf{91.6} & --- & \textbf{78.3} & \textbf{97.6} & \textbf{94.9} & --- & \textbf{86.8} & \textbf{98.1} & \textbf{96.9} & --- & \textbf{94.5} & \textbf{97.4} & \textbf{95.7} & --- & \textbf{89.9} \\

& Gray & 91.3 & 76.8 & --- & 51.1 & 94.0 & 89.1 & --- & 76.4 & 97.3 & 93.5 & --- & 85.6 & 97.9 & 96.0 & --- & 92.8 & 97.2 & 95.2 & --- & 89.4 \\
\hline

\multirow{2}{*}{\textbf{Winter}}
& RGB  & \textbf{52.0} & \textbf{71.1} & \textbf{47.3} & --- & \textbf{70.5} & \textbf{86.9} & \textbf{65.6} & --- & \textbf{86.9} & \textbf{93.9} & \textbf{84.5} & --- & \textbf{93.6} & \textbf{97.4} & \textbf{92.4} & --- & 89.6 & \textbf{96.7} & 88.6 & --- \\

& Gray & 43.6 & 62.9 & 39.8 & --- & 67.5 & 84.1 & 64.3 & --- & 86.6 & 93.7 & 83.9 & --- & 92.3 & \textbf{97.4} & 90.5 & --- & \textbf{90.4} & \textbf{96.7} & \textbf{88.7} & --- \\
\hline
\hline
\multicolumn{2}{|c|}{\textbf{Avg.\ $\Delta$R@1 (RGB$-$Gray)}} &
\multicolumn{4}{c|}{7.6} &
\multicolumn{4}{c|}{2.0} &
\multicolumn{4}{c|}{0.6} &
\multicolumn{4}{c|}{1.0} &
\multicolumn{4}{c|}{0.3} \\
\hline
\end{tabularx}
\vspace*{-1em}
\end{table*}
}
\setlength{\tablew}{\textwidth}
\setlength{\firstcolw}{1.3cm}
\newcommand\notsotiny{\fontsize{6.5pt}{7.5pt}\selectfont}

\begin{table*}[!ht]
\notsotiny
\centering
\caption{Recall@1 across Oxford RobotCar weather pairs, evaluated on RGB and Gray using \textbf{off-the-shelf Models}.}
\label{tab:ORC_OTS}
\setlength{\tabcolsep}{1.9pt}
\renewcommand{\arraystretch}{1.0}
\begin{tabularx}{\tablew}{|p{\firstcolw}|c|YYYYYY|YYYYYY|YYYYYY|YYYYYY|YYYYYY|}
\hline
& \textbf{Model} $\rightarrow$ &
\multicolumn{6}{c|}{\textbf{EigenPlaces}} &
\multicolumn{6}{c|}{\textbf{MixVPR}} &
\multicolumn{6}{c|}{\textbf{SALAD}} &
\multicolumn{6}{c|}{\textbf{CricaVPR}} &
\multicolumn{6}{c|}{\textbf{BoQ}} \\
\cline{2-32}
\makecell[l]{\textbf{Reference} $\downarrow$} &
\makecell[b]{\textbf{Query} $\rightarrow$ \\[0.35em] \textbf{Eval. Mode} $\downarrow$} &
\rot{\textbf{Sun}} & \rot{\textbf{Dusk}} & \rot{\textbf{Overcast}} & \rot{\textbf{Rain}} & \rot{\textbf{Night}} & \rot{\textbf{Snow}} &
\rot{\textbf{Sun}} & \rot{\textbf{Dusk}} & \rot{\textbf{Overcast}} & \rot{\textbf{Rain}} & \rot{\textbf{Night}} & \rot{\textbf{Snow}} &
\rot{\textbf{Sun}} & \rot{\textbf{Dusk}} & \rot{\textbf{Overcast}} & \rot{\textbf{Rain}} & \rot{\textbf{Night}} & \rot{\textbf{Snow}} &
\rot{\textbf{Sun}} & \rot{\textbf{Dusk}} & \rot{\textbf{Overcast}} & \rot{\textbf{Rain}} & \rot{\textbf{Night}} & \rot{\textbf{Snow}} &
\rot{\textbf{Sun}} & \rot{\textbf{Dusk}} & \rot{\textbf{Overcast}} & \rot{\textbf{Rain}} & \rot{\textbf{Night}} & \rot{\textbf{Snow}} \\
\hline\hline
\multirow{2}{*}{\textbf{Sun}}
& RGB  & --- & \textbf{59.3} & \textbf{96.7} & \textbf{94.8} & \textbf{44.8} & \textbf{87.4} & --- & \textbf{64.3} & \textbf{97.4} & \textbf{96.9} & 66.3 & \textbf{90.4} & --- & \textbf{71.4} & \textbf{96.6} & \textbf{95.8} & \textbf{83.5} & \textbf{89.4} & --- & \textbf{72.0} & \textbf{97.1} & 96.6 & \textbf{81.5} & \textbf{89.3} & --- & \textbf{77.5} & \textbf{98.0} & \textbf{97.4} & \textbf{89.8} & \textbf{90.9} \\

& Gray & --- & 51.4 & 96.2 & 93.4 & 42.5 & 86.6 & --- & 57.9 & 95.4 & 94.7 & \textbf{68.3} & 85.1 & --- & 67.9 & 96.1 & 95.1 & 82.6 & 86.7 & --- & 68.6 & 97.1 & \textbf{96.6} & 80.5 & 87.9 & --- & 75.4 & 97.8 & 97.1 & 89.4 & 89.0 \\
\hline

\multirow{2}{*}{\textbf{Dusk}}
& RGB  & \textbf{54.4} & --- & \textbf{55.5} & \textbf{50.4} & \textbf{37.1} & \textbf{58.0} & \textbf{58.2} & --- & \textbf{57.7} & \textbf{52.6} & \textbf{51.9} & \textbf{59.6} & \textbf{66.7} & --- & \textbf{66.5} & \textbf{62.3} & \textbf{71.1} & \textbf{67.1} & \textbf{66.3} & --- & \textbf{67.4} & \textbf{62.7} & \textbf{69.2} & \textbf{67.7} & \textbf{75.0} & --- & \textbf{75.9} & \textbf{68.3} & \textbf{75.8} & \textbf{72.6} \\

& Gray & 47.3 & --- & 49.0 & 43.5 & 30.3 & 49.3 & 53.3 & --- & 53.7 & 49.0 & 48.1 & 56.4 & 64.4 & --- & 64.3 & 58.3 & 65.8 & 63.4 & 62.7 & --- & 65.6 & 60.6 & 65.2 & 65.5 & 73.7 & --- & 73.5 & 65.0 & 71.7 & 70.3 \\
\hline

\multirow{2}{*}{\textbf{Overcast}}
& RGB  & 96.5 & \textbf{60.0} & --- & \textbf{94.2} & \textbf{50.5} & \textbf{89.7} & \textbf{96.7} & \textbf{67.0} & --- & \textbf{95.8} & 72.5 & \textbf{91.3} & 96.2 & \textbf{72.4} & --- & 94.8 & 84.7 & \textbf{90.3} & 96.7 & \textbf{72.9} & --- & \textbf{95.9} & \textbf{83.0} & \textbf{91.0} & \textbf{97.4} & \textbf{80.2} & --- & \textbf{96.5} & \textbf{91.6} & \textbf{92.8} \\

& Gray & \textbf{96.6} & 53.2 & --- & 93.7 & 47.7 & 89.2 & 96.0 & 60.1 & --- & 95.1 & \textbf{73.7} & 87.4 & \textbf{96.4} & 69.0 & --- & \textbf{95.1} & \textbf{85.5} & 89.4 & \textbf{96.7} & 71.7 & --- & 95.7 & 81.5 & 90.7 & 97.2 & 78.5 & --- & 96.4 & 90.8 & 91.5 \\
\hline

\multirow{2}{*}{\textbf{Rain}}
& RGB  & \textbf{95.5} & \textbf{55.1} & \textbf{95.0} & --- & \textbf{46.8} & \textbf{85.8} & \textbf{95.8} & \textbf{57.6} & \textbf{95.2} & --- & \textbf{65.7} & \textbf{85.8} & 94.7 & \textbf{64.0} & \textbf{95.5} & --- & \textbf{83.1} & \textbf{87.2} & 96.2 & \textbf{67.7} & 95.0 & --- & \textbf{81.5} & \textbf{86.4} & 96.8 & \textbf{71.2} & 96.9 & --- & \textbf{88.8} & \textbf{89.3} \\

& Gray & 94.0 & 47.0 & 93.7 & --- & 40.4 & 84.4 & 94.8 & 52.1 & 94.4 & --- & 64.9 & 82.9 & \textbf{95.7} & 61.4 & 95.3 & --- & 82.1 & 85.7 & \textbf{96.7} & 65.0 & \textbf{96.0} & --- & 80.4 & 86.3 & \textbf{96.9} & 67.7 & \textbf{97.3} & --- & 87.0 & 88.2 \\
\hline

\multirow{2}{*}{\textbf{Night}}
& RGB  & \textbf{64.8} & \textbf{47.8} & \textbf{70.2} & \textbf{61.7} & --- & \textbf{61.2} & 71.1 & \textbf{60.2} & 74.4 & 70.3 & --- & \textbf{70.5} & 80.8 & \textbf{71.8} & 80.2 & 80.0 & --- & 78.6 & \textbf{81.8} & \textbf{70.6} & 82.9 & 80.8 & --- & 77.5 & \textbf{88.9} & \textbf{77.2} & \textbf{89.6} & \textbf{87.1} & --- & 84.0 \\

& Gray & 61.2 & 42.6 & 64.8 & 55.7 & --- & 56.8 & \textbf{71.1} & 53.2 & \textbf{74.4} & \textbf{70.6} & --- & 65.4 & \textbf{83.8} & 68.4 & \textbf{84.2} & \textbf{81.1} & --- & \textbf{79.3} & 81.3 & 69.0 & \textbf{84.7} & \textbf{81.4} & --- & \textbf{78.7} & 88.0 & 76.2 & 88.8 & 86.8 & --- & \textbf{85.2} \\
\hline

\multirow{2}{*}{\textbf{Snow}}
& RGB  & \textbf{89.9} & \textbf{66.7} & \textbf{93.0} & \textbf{88.4} & \textbf{52.3} & --- & \textbf{92.5} & \textbf{71.5} & \textbf{94.4} & \textbf{90.4} & \textbf{71.0} & --- & \textbf{92.0} & \textbf{72.7} & 93.4 & \textbf{89.9} & 83.7 & --- & \textbf{92.9} & \textbf{77.5} & \textbf{94.4} & \textbf{90.8} & \textbf{83.8} & --- & \textbf{93.7} & \textbf{79.5} & \textbf{95.5} & \textbf{92.0} & 88.8 & --- \\

& Gray & 89.2 & 57.4 & 92.3 & 87.2 & 50.6 & --- & 88.8 & 65.7 & 92.0 & 86.9 & 65.2 & --- & 90.2 & 71.4 & \textbf{93.6} & 89.5 & \textbf{83.8} & --- & 91.6 & 73.8 & 94.3 & 90.5 & 82.2 & --- & 93.2 & 78.1 & 95.1 & 91.5 & \textbf{89.3} & --- \\
\hline  
\hline
\multicolumn{2}{|c|}{\textbf{Avg.\ $\Delta$R@1 (RGB$-$Gray)}} &
\multicolumn{6}{c|}{3.9} &
\multicolumn{6}{c|}{2.9} &
\multicolumn{6}{c|}{1.0} &
\multicolumn{6}{c|}{1.0} &
\multicolumn{6}{c|}{1.1} \\
\hline
\end{tabularx}
\vspace*{-1em}
\end{table*}










\begin{table}[!ht]
\notsotiny
\centering
\caption{Recall@1 across datasets, evaluated on RGB and Gray using \textbf{off-the-shelf Models}.}
\label{tab:rest_OTS}
\setlength{\tabcolsep}{2.9pt}
\renewcommand{\arraystretch}{1.0}
\begin{tabular}{|p{\firstcolw}|c|cccc|c|}
\hline
\textbf{Model} &
\textbf{Eval. Mode} &
\makecell{\textbf{Pitts250k} \\ \textbf{(Test)}} &
\textbf{SPED} &
\textbf{GLDv2} &
\textbf{Tokyo247} &
\makecell{\textbf{Avg.$\Delta$R@1} \\ \textbf{(RGB-Gray)}} \\
\hline\hline
\multirow{2}{*}{\textbf{EigenPlaces}}
& RGB  & \textbf{95.1} & \textbf{78.2} & \textbf{53.6} & \textbf{80.6} & \multirow{2}{*}{3.4} \\
& Gray & 94.4 & 75.4 & 51.0 & 73.0 & \\
\hline
\multirow{2}{*}{\textbf{MixVPR}}
& RGB  & \textbf{96.0} & 84.7 & \textbf{59.5} & \textbf{86.4} & \multirow{2}{*}{5.0} \\
& Gray & 93.8 & \textbf{86.5} & 53.8 & 72.4 & \\
\hline
\multirow{2}{*}{\textbf{SALAD}}
& RGB  & \textbf{97.4} & \textbf{92.1} & \textbf{71.7} & \textbf{94.6} & \multirow{2}{*}{2.1} \\
& Gray & 96.7 & 90.9 & 67.7 & 92.1 & \\
\hline
\multirow{2}{*}{\textbf{CricaVPR}}
& RGB  & \textbf{96.2} & \textbf{92.9} & \textbf{70.2} & \textbf{93.0} & \multirow{2}{*}{1.3} \\
& Gray & 95.6 & \textbf{92.9} & 67.7 & 90.8 & \\
\hline
\multirow{2}{*}{\textbf{BoQ}}
& RGB  & \textbf{97.6} & \textbf{92.6} & \textbf{74.2} & \textbf{96.5} & \multirow{2}{*}{1.2} \\
& Gray & \textbf{97.6} & 91.4 & 71.0 & 96.2 & \\
\hline
\end{tabular}
\vspace{-2em}
\end{table}

\textbf{Impact of Color in Vision Tasks.} Color utility is largely task and data-dependent in both human and machine vision. In human vision, color supports object recognition \cite{BRAMAO2011244}, face recognition \cite{doi:10.1068/p3376,Brosseau20042020}, and edge classification \cite{10.1371/journal.pcbi.1007398}, yet scene recognition studies show it does not consistently improve accuracy, instead contributing to memory retrieval and decision latency \cite{10.1167/8.16.12,NIJBOER2008741}. Spatial structure primarily drives real-world visual search \cite{Ehinger2008}, and early scene categorization is largely unaffected by color absence \cite{DELORME20002187}. Late-sighted individuals exhibit over-reliance on color \cite{doi:10.1126/science.adk9587}, implicating luminance-based structure as the foundation for robust learning. In computer vision, color aids object-centric recognition \cite{singh2020colours,doi:10.1126/science.adk9587} but has little effect on face recognition under grayscale \cite{10887263}. Luminance-based structure is more robust to color space changes and advantageous in texture-dominant settings \cite{10.1007/978-3-030-31332-6_27}. At the representation level, color sensitivity was found to concentrate in early layers of CNNs and diminishing as features become class-specific \cite{depthcol}. Surrounding object-centric data, ColorSense \cite{10992338} finds that performance gaps between color-contrast difficulty groups persist regardless of architecture or augmentation, with grayscale only partially reducing bias, and \cite{9607588} highlight that RGB-trained models degrade significantly under chromatic shifts, yet color's role in VPR for robotics and mobile deployment remains underexplored. Unlike such tasks, VPR requires globally consistent representations across severe appearance and structural change. Existing evaluations address appearance variation, viewpoint change and degradation from architectural and loss perspectives \cite{10938388,8461051,zaffar2021vpr,9336674}, with limited analysis of color-induced variation. Under illumination, weather and seasonal shifts, color is unstable and can harm retrieval if models lack invariance. This raises the question of whether structural and texture cues alone suffice, and whether grayscale representations offer lower-variance, robust alternatives with reduced overhead.

\setlength{\tablew}{\textwidth}
\setlength{\firstcolw}{1.8cm}
{
\begin{table*}[!ht]
\footnotesize
\centering
\caption{Recall@1 across Nordland season pairs with MixVPR trained \& evaluated per modality, \textbf{RGB ImageNet pretrained}.}
\label{tab:nordland_RGBP}
\setlength{\tabcolsep}{4.0pt}
\renewcommand{\arraystretch}{1.0}
\begin{tabular}{|p{\firstcolw}|c|cccc|cccc|cccc|}
\hline
& \textbf{Model} $\rightarrow$ &
\multicolumn{4}{c|}{\textbf{MixVPR -- ResNet18}} &
\multicolumn{4}{c|}{\textbf{MixVPR -- ResNet34}} &
\multicolumn{4}{c|}{\textbf{MixVPR -- ResNet50}} \\
\cline{2-14}
\makecell[l]{\textbf{Reference} $\downarrow$} &
\makecell[b]{\textbf{Query} $\rightarrow$ \\[0.35em] \textbf{Mode} $\downarrow$} &
\rot{\textbf{Fall}} & \rot{\textbf{Spring}} & \rot{\textbf{Summer}} & \rot{\textbf{Winter}} &
\rot{\textbf{Fall}} & \rot{\textbf{Spring}} & \rot{\textbf{Summer}} & \rot{\textbf{Winter}} &
\rot{\textbf{Fall}} & \rot{\textbf{Spring}} & \rot{\textbf{Summer}} & \rot{\textbf{Winter}} \\
\hline\hline
\multirow{2}{*}{\textbf{Fall}}
& RGB  & --- & \textbf{94.3} & \textbf{94.7} & \textbf{75.3} & --- & \textbf{91.8} & \textbf{94.5} & 71.5 & --- & \textbf{95.6} & \textbf{95.5} & 82.7 \\
& Gray & --- & 92.7 & 94.3 & 74.9 & --- & 91.6 & 93.6 & \textbf{74.1} & --- & 94.9 & 95.0 & \textbf{83.5} \\
\hline

\multirow{2}{*}{\textbf{Spring}}
& RGB  & \textbf{87.9} & --- & \textbf{80.6} & 84.0 & \textbf{83.7} & --- & 77.6 & 82.0 & \textbf{89.9} & --- & \textbf{84.4} & 89.7 \\
& Gray & 85.5 & --- & 78.6 & \textbf{86.7} & 83.5 & --- & \textbf{78.5} & \textbf{86.2} & 89.2 & --- & 84.0 & \textbf{92.9} \\
\hline

\multirow{2}{*}{\textbf{Summer}}
& RGB  & \textbf{94.8} & \textbf{88.8} & --- & \textbf{73.0} & \textbf{95.0} & \textbf{88.8} & --- & 70.5 & \textbf{96.1} & \textbf{91.3} & --- & 80.4 \\
& Gray & 94.1 & 86.9 & --- & 72.9 & 93.7 & 87.6 & --- & \textbf{70.6} & 95.2 & 91.1 & --- & \textbf{80.8} \\
\hline

\multirow{2}{*}{\textbf{Winter}}
& RGB  & 64.7 & 82.9 & 60.9 & --- & 59.0 & 78.6 & 56.7 & --- & 74.9 & 88.5 & 70.3 & --- \\
& Gray & \textbf{66.8} & \textbf{83.8} & \textbf{63.6} & --- & \textbf{63.3} & \textbf{82.1} & \textbf{61.8} & --- & \textbf{76.2} & \textbf{89.8} & \textbf{73.0} & --- \\
\hline\hline
\multicolumn{2}{|c|}{\textbf{Avg.\ $\Delta$R@1 (RGB$-$Gray)}} &
\multicolumn{4}{c|}{0.1} &
\multicolumn{4}{c|}{-1.4} &
\multicolumn{4}{c|}{-0.5} \\
\hline
\end{tabular}
\vspace*{-1em}
\end{table*}
}
\setlength{\tablew}{\textwidth}
\setlength{\firstcolw}{1.5cm}
\begin{table*}[!ht]
\footnotesize
\centering
\caption{Recall@1 across Oxford RobotCar weather pairs with MixVPR trained \& evaluated per modality, \textbf{RGB ImageNet pretrained}.}
\label{tab:ORC_RGBP}
\setlength{\tabcolsep}{1.9pt}
\renewcommand{\arraystretch}{1.0}
\begin{tabularx}{\tablew}{|p{\firstcolw}|c|YYYYYY|YYYYYY|YYYYYY|}
\hline
& \textbf{Model} $\rightarrow$ &
\multicolumn{6}{c|}{\textbf{MixVPR -- ResNet18}} &
\multicolumn{6}{c|}{\textbf{MixVPR -- ResNet34}} &
\multicolumn{6}{c|}{\textbf{MixVPR -- ResNet50}} \\
\cline{2-20}
\makecell[l]{\textbf{Reference} $\downarrow$} &
\makecell[b]{\textbf{Query} $\rightarrow$ \\[0.35em] \textbf{Mode} $\downarrow$} &
\rot{\textbf{Sun}} & \rot{\textbf{Dusk}} & \rot{\textbf{Overcast}} & \rot{\textbf{Rain}} & \rot{\textbf{Night}} & \rot{\textbf{Snow}} &
\rot{\textbf{Sun}} & \rot{\textbf{Dusk}} & \rot{\textbf{Overcast}} & \rot{\textbf{Rain}} & \rot{\textbf{Night}} & \rot{\textbf{Snow}} &
\rot{\textbf{Sun}} & \rot{\textbf{Dusk}} & \rot{\textbf{Overcast}} & \rot{\textbf{Rain}} & \rot{\textbf{Night}} & \rot{\textbf{Snow}} \\
\hline\hline
\multirow{2}{*}{\textbf{Sun}}\textbf{}
& RGB  & --- & 57.9 & 96.5 & \textbf{95.3} & 50.0 & 87.3 & --- & 59.2 & 96.8 & 95.0 & 58.4 & 87.9 & --- & 65.8 & \textbf{97.5} & 96.5 & 67.5 & 90.4 \\
& Gray & --- & \textbf{60.3} & \textbf{97.0} & 95.2 & \textbf{65.4} & \textbf{88.5} & --- & \textbf{63.4} & \textbf{97.0} & \textbf{96.2} & \textbf{71.3} & \textbf{89.0} & --- & \textbf{69.0} & 97.4 & \textbf{96.7} & \textbf{78.9} & \textbf{90.6} \\
\hline

\multirow{2}{*}{\textbf{Dusk}}
& RGB  & \textbf{54.1} & --- & 53.7 & 51.1 & 38.4 & 58.0 & 55.7 & --- & 56.1 & \textbf{52.6} & 47.6 & 57.9 & 55.7 & --- & 56.5 & 53.1 & 53.1 & 59.8 \\
& Gray & \textbf{54.1} & --- & \textbf{55.6} & \textbf{51.2} & \textbf{52.6} & \textbf{58.9} & \textbf{57.0} & --- & \textbf{58.0} & 52.0 & \textbf{54.9} & \textbf{58.6} & \textbf{60.7} & --- & \textbf{61.0} & \textbf{54.8} & \textbf{61.2} & \textbf{62.7} \\
\hline

\multirow{2}{*}{\textbf{Overcast}}
& RGB  & 96.4 & 59.4 & --- & 94.6 & 52.9 & 89.4 & \textbf{96.2} & 60.5 & --- & 94.2 & 61.9 & 89.7 & 96.9 & 65.6 & --- & 96.0 & 69.2 & 91.7 \\
& Gray & \textbf{96.6} & \textbf{61.4} & --- & \textbf{94.9} & \textbf{69.1} & \textbf{90.9} & 95.9 & \textbf{63.6} & --- & \textbf{95.1} & \textbf{74.3} & \textbf{91.9} & \textbf{97.3} & \textbf{71.0} & --- & \textbf{96.4} & \textbf{83.2} & \textbf{93.0} \\
\hline

\multirow{2}{*}{\textbf{Rain}}
& RGB  & 93.4 & \textbf{53.7} & 93.0 & --- & 46.9 & 82.4 & 94.1 & 54.2 & 93.6 & --- & 53.9 & 82.8 & 95.4 & 56.1 & 94.8 & --- & 62.0 & 85.8 \\
& Gray & \textbf{96.0} & 53.2 & \textbf{94.7} & --- & \textbf{61.5} & \textbf{84.6} & \textbf{95.6} & \textbf{56.6} & \textbf{95.4} & --- & \textbf{68.2} & \textbf{85.9} & \textbf{96.9} & \textbf{60.9} & \textbf{96.6} & --- & \textbf{74.2} & \textbf{87.2} \\
\hline

\multirow{2}{*}{\textbf{Night}}
& RGB  & 55.0 & 49.6 & 57.7 & 53.9 & --- & 54.1 & 61.1 & 53.2 & 65.9 & 58.2 & --- & 60.8 & 68.0 & 57.7 & 71.4 & 66.3 & --- & 66.7 \\
& Gray & \textbf{70.0} & \textbf{56.1} & \textbf{73.2} & \textbf{67.3} & --- & \textbf{67.2} & \textbf{75.0} & \textbf{60.2} & \textbf{77.8} & \textbf{72.0} & --- & \textbf{72.5} & \textbf{78.7} & \textbf{66.6} & \textbf{82.7} & \textbf{77.8} & --- & \textbf{78.3} \\
\hline

\multirow{2}{*}{\textbf{Snow}}
& RGB  & 89.2 & 67.3 & 92.6 & 87.4 & 53.1 & --- & 90.7 & 67.5 & 92.8 & 86.8 & 60.8 & --- & \textbf{92.8} & 71.8 & 94.8 & 90.8 & 69.7 & --- \\
& Gray & \textbf{91.2} & \textbf{68.9} & \textbf{93.2} & \textbf{88.2} & \textbf{67.2} & --- & \textbf{91.4} & \textbf{69.3} & \textbf{94.0} & \textbf{88.7} & \textbf{72.8} & --- & 92.3 & \textbf{75.7} & \textbf{95.1} & \textbf{91.3} & \textbf{81.3} & --- \\
\hline\hline
\multicolumn{2}{|c|}{\textbf{Avg.\ $\Delta$R@1 (RGB$-$Gray)}} &
\multicolumn{6}{c|}{-5.3} &
\multicolumn{6}{c|}{-4.9} &
\multicolumn{6}{c|}{-5.0} \\
\hline
\end{tabularx}
\vspace*{-1em}
\end{table*}
\begin{table}[!ht]
\notsotiny
\centering
\caption{Recall@1 across datasets with MixVPR trained \& evaluated per modality, \textbf{RGB ImageNet pretrained}.}
\label{tab:rest_RGBP}
\setlength{\firstcolw}{1.3cm}
\setlength{\tabcolsep}{2.9pt}
\renewcommand{\arraystretch}{1.0}
\begin{tabular}{|p{\firstcolw}|c|cccc|c|}
\hline
\textbf{Model} &
\textbf{Mode} &
\makecell{\textbf{Pitts250k} \\ \textbf{(Test)}} &
\textbf{SPED} &
\textbf{GLDv2} &
\textbf{Tokyo247} &
\makecell{\textbf{Avg.$\Delta$R@1} \\ \textbf{(RGB-Gray)}} \\
\hline\hline
\multirow{2}{*}{\makecell[l]{\textbf{MixVPR} \\ \textbf{(ResNet18)}}}
& RGB  & \textbf{94.0} & 82.2 & \textbf{51.1} & \textbf{74.0}  & \multirow{2}{*}{0.2} \\
& Gray & 93.6 & \textbf{84.2} & 49.4 & 73.3 & \\
\hline
\multirow{2}{*}{\makecell[l]{\textbf{MixVPR} \\ \textbf{(ResNet34)}}}
& RGB  & 94.8 & 82.5 & \textbf{53.3} & \textbf{79.4}  & \multirow{2}{*}{1.7} \\
& Gray & \textbf{94.9} & \textbf{84.7} & 49.7 & 74.0 & \\
\hline
\multirow{2}{*}{\makecell[l]{\textbf{MixVPR} \\ \textbf{(ResNet50)}}}
& RGB  & \textbf{96.4} & 83.7 & \textbf{58.9} & \textbf{86.4}  & \multirow{2}{*}{0.1} \\
& Gray & 96.0 & \textbf{85.3} & 57.1 & \textbf{86.4} & \\
\hline
\end{tabular}
\vspace{-2em}
\end{table}

\section{EXPERIMENTAL METHODOLOGY} 
\label{sec:expsetup}
We evaluate the impact of color on VPR across architectures and training regimes, structured into setups shown in \fig{fig:blockdiag}.

\textbf{Setup 1: Off-the-shelf.} Existing VPR models (\tab{tab:models}) are evaluated without modification, covering ResNet-based (EigenPlaces, MixVPR) and DINOv2-based methods (SALAD, CricaVPR, BoQ), all RGB pretrained and VPR-trained. Each is evaluated on both RGB and grayscale benchmarks. For grayscale, the single image channel is replicated across three to match the model input format.

\textbf{Setups 2 \& 3: Controlled Training.} We isolate MixVPR for its simplicity and lightweight design. DINOv2-based models are excluded as LVD-142M is unavailable for replication. ResNet backbones are pretrained on RGB and grayscale ImageNet from scratch, then fine-tuned and evaluated on either modality, yielding four variants across ResNet18, 34 and 50. All experiments share the same seed, hardware and hyperparameters. The full model is kept unfrozen during VPR fine-tuning. RGB input weights are averaged to one channel for grayscale fine-tuning, and grayscale input weights are replicated across three for RGB.

\textbf{Setup 4: VPR Training from Scratch.} MixVPR is trained directly on VPR data without ImageNet pretraining across both modalities and all three backbone sizes, isolating color's role in the VPR objective and examining whether GSV-Cities provides sufficient color invariance for generalization.

\textbf{Datasets.} Models are evaluated on six benchmarks covering seasonal, weather, illumination and viewpoint change. 


\setlength{\tablew}{\textwidth}
\setlength{\firstcolw}{1.8cm}
{
\begin{table*}[!ht]
\footnotesize
\centering
\caption{Recall@1 across Nordland season pairs with MixVPR trained \& evaluated per modality, \textbf{Gray ImageNet pretrained}.}
\label{tab:nordland_GrayP}
\setlength{\tabcolsep}{4.0pt}
\renewcommand{\arraystretch}{1.0}
\begin{tabular}{|p{\firstcolw}|c|cccc|cccc|cccc|}
\hline
& \textbf{Model} $\rightarrow$ &
\multicolumn{4}{c|}{\textbf{MixVPR -- ResNet18}} &
\multicolumn{4}{c|}{\textbf{MixVPR -- ResNet34}} &
\multicolumn{4}{c|}{\textbf{MixVPR -- ResNet50}} \\
\cline{2-14}
\makecell[l]{\textbf{Reference} $\downarrow$} &
\makecell[b]{\textbf{Query} $\rightarrow$ \\[0.35em] \textbf{Mode} $\downarrow$} &
\rot{\textbf{Fall}} & \rot{\textbf{Spring}} & \rot{\textbf{Summer}} & \rot{\textbf{Winter}} &
\rot{\textbf{Fall}} & \rot{\textbf{Spring}} & \rot{\textbf{Summer}} & \rot{\textbf{Winter}} &
\rot{\textbf{Fall}} & \rot{\textbf{Spring}} & \rot{\textbf{Summer}} & \rot{\textbf{Winter}} \\
\hline\hline
\multirow{2}{*}{\textbf{Fall}}
& RGB  & --- & \textbf{94.2} & \textbf{94.6} & 75.9 & --- & 91.8 & \textbf{94.4} & 73.2 & --- & 95.1 & \textbf{95.8} & \textbf{82.6} \\
& Gray & --- & 93.4 & 94.0 & \textbf{76.4} & --- & \textbf{92.8} & 93.8 & \textbf{73.6} & --- & \textbf{95.4} & 95.4 & \textbf{82.6} \\
\hline
\multirow{2}{*}{\textbf{Spring}}
& RGB  & 86.9 & --- & 80.1 & 85.6 & 83.1 & --- & 78.6 & 82.0 & 88.0 & --- & 82.5 & 89.7 \\
& Gray & \textbf{89.9} & --- & \textbf{83.6} & \textbf{88.2} & \textbf{85.6} & --- & \textbf{80.4} & \textbf{85.8} & \textbf{90.5} & --- & \textbf{85.4} & \textbf{91.6} \\
\hline
\multirow{2}{*}{\textbf{Summer}}
& RGB  & \textbf{95.1} & \textbf{89.3} & --- & 73.0 & \textbf{93.9} & 87.9 & --- & 70.5 & \textbf{96.1} & 91.0 & --- & 78.3 \\
& Gray & 94.3 & 88.5 & --- & \textbf{73.5} & 93.8 & \textbf{88.7} & --- & \textbf{71.7} & 95.7 & \textbf{91.3} & --- & \textbf{78.9} \\
\hline
\multirow{2}{*}{\textbf{Winter}}
& RGB  & 67.8 & 84.1 & 64.2 & --- & 63.0 & 81.3 & 61.0 & --- & 70.9 & 87.5 & 66.4 & --- \\
& Gray & \textbf{69.7} & \textbf{84.8} & \textbf{66.6} & --- & \textbf{65.0} & \textbf{84.3} & \textbf{62.3} & --- & \textbf{75.5} & \textbf{89.8} & \textbf{71.0} & --- \\
\hline\hline
\multicolumn{2}{|c|}{\textbf{Avg.\ $\Delta$R@1 (RGB$-$Gray)}} &
\multicolumn{4}{c|}{-1.0} &
\multicolumn{4}{c|}{-1.4} &
\multicolumn{4}{c|}{-1.6} \\
\hline
\end{tabular}
\vspace*{-1em}
\end{table*}
}
\setlength{\tablew}{\textwidth}
\setlength{\firstcolw}{1.5cm}
\begin{table*}[!ht]
\footnotesize
\centering
\caption{Recall@1 across Oxford RobotCar weather pairs with MixVPR trained \& evaluated per modality, \textbf{Gray ImageNet pretrained}.}
\label{tab:ORC_GrayP}
\setlength{\tabcolsep}{1.9pt}
\renewcommand{\arraystretch}{1.0}
\begin{tabularx}{\tablew}{|p{\firstcolw}|c|YYYYYY|YYYYYY|YYYYYY|}
\hline
& \textbf{Model} $\rightarrow$ &
\multicolumn{6}{c|}{\textbf{MixVPR -- ResNet18}} &
\multicolumn{6}{c|}{\textbf{MixVPR -- ResNet34}} &
\multicolumn{6}{c|}{\textbf{MixVPR -- ResNet50}} \\
\cline{2-20}
\makecell[l]{\textbf{Reference} $\downarrow$} &
\makecell[b]{\textbf{Query} $\rightarrow$ \\[0.35em] \textbf{Mode} $\downarrow$} &
\rot{\textbf{Sun}} & \rot{\textbf{Dusk}} & \rot{\textbf{Overcast}} & \rot{\textbf{Rain}} & \rot{\textbf{Night}} & \rot{\textbf{Snow}} &
\rot{\textbf{Sun}} & \rot{\textbf{Dusk}} & \rot{\textbf{Overcast}} & \rot{\textbf{Rain}} & \rot{\textbf{Night}} & \rot{\textbf{Snow}} &
\rot{\textbf{Sun}} & \rot{\textbf{Dusk}} & \rot{\textbf{Overcast}} & \rot{\textbf{Rain}} & \rot{\textbf{Night}} & \rot{\textbf{Snow}} \\
\hline\hline
\multirow{2}{*}{\textbf{Sun}}
& RGB  & --- & 58.7 & 96.9 & 95.8 & 49.4 & 87.9 & --- & 59.3 & \textbf{97.1} & 95.0 & 54.9 & \textbf{88.8} & --- & 64.9 & \textbf{97.4} & \textbf{96.7} & 68.4 & 89.5 \\
& Gray & --- & \textbf{60.0} & \textbf{97.0} & \textbf{96.0} & \textbf{67.2} & \textbf{88.9} & --- & \textbf{60.9} & 97.0 & \textbf{96.0} & \textbf{72.5} & 88.4 & --- & \textbf{66.4} & 97.2 & \textbf{96.7} & \textbf{79.8} & \textbf{89.9} \\
\hline

\multirow{2}{*}{\textbf{Dusk}}
& RGB  & 53.8 & --- & 54.5 & \textbf{51.8} & 40.7 & 57.2 & \textbf{53.8} & --- & 52.0 & \textbf{50.4} & 45.3 & \textbf{59.3} & 56.9 & --- & 56.4 & 53.5 & 54.0 & 59.6 \\
& Gray & \textbf{54.7} & --- & \textbf{55.1} & 50.8 & \textbf{54.0} & \textbf{58.0} & 53.3 & --- & \textbf{54.3} & 49.8 & \textbf{53.9} & 57.6 & \textbf{58.5} & --- & \textbf{59.6} & \textbf{54.3} & \textbf{60.2} & \textbf{61.0} \\
\hline

\multirow{2}{*}{\textbf{Overcast}}
& RGB  & 96.4 & 60.2 & --- & 95.1 & 53.4 & 90.3 & 96.4 & \textbf{61.3} & --- & 95.7 & 60.4 & 90.5 & \textbf{97.3} & 66.5 & --- & 96.3 & 74.2 & 92.5 \\
& Gray & \textbf{96.8} & \textbf{63.6} & --- & \textbf{96.2} & \textbf{74.0} & \textbf{91.0} & \textbf{96.6} & 62.1 & --- & \textbf{96.1} & \textbf{77.2} & \textbf{91.7} & \textbf{97.3} & \textbf{68.3} & --- & \textbf{96.9} & \textbf{83.4} & \textbf{92.8} \\
\hline

\multirow{2}{*}{\textbf{Rain}}
& RGB  & 94.9 & 52.0 & 93.8 & --- & 46.5 & 82.9 & 94.7 & \textbf{54.6} & 94.1 & --- & 51.5 & 84.1 & 96.7 & 58.1 & 95.9 & --- & 64.7 & 87.1 \\
& Gray & \textbf{95.5} & \textbf{54.8} & \textbf{95.2} & --- & \textbf{63.8} & \textbf{85.0} & \textbf{95.8} & 54.4 & \textbf{95.4} & --- & \textbf{68.9} & \textbf{85.5} & \textbf{97.0} & \textbf{61.2} & \textbf{96.5} & --- & \textbf{75.9} & \textbf{87.5} \\
\hline

\multirow{2}{*}{\textbf{Night}}
& RGB  & 54.8 & 47.5 & 60.5 & 56.0 & --- & 55.0 & 57.8 & 53.3 & 62.2 & 57.8 & --- & 59.1 & 70.1 & 60.9 & 74.5 & 67.4 & --- & 68.2 \\
& Gray & \textbf{72.2} & \textbf{55.9} & \textbf{77.1} & \textbf{69.1} & --- & \textbf{69.5} & \textbf{72.9} & \textbf{58.2} & \textbf{77.7} & \textbf{70.4} & --- & \textbf{70.8} & \textbf{79.2} & \textbf{66.7} & \textbf{83.7} & \textbf{77.1} & --- & \textbf{77.8} \\
\hline

\multirow{2}{*}{\textbf{Snow}}
& RGB  & 90.3 & 67.2 & 93.6 & 86.8 & 55.3 & --- & 90.3 & \textbf{67.7} & 93.0 & 87.7 & 59.7 & --- & 92.7 & 72.0 & 95.1 & 90.6 & 71.4 & --- \\
& Gray & \textbf{90.7} & \textbf{69.0} & \textbf{94.0} & \textbf{88.1} & \textbf{72.8} & --- & \textbf{91.8} & 67.0 & \textbf{93.2} & \textbf{88.5} & \textbf{72.5} & --- & \textbf{93.8} & \textbf{74.0} & \textbf{95.6} & \textbf{90.7} & \textbf{80.3} & --- \\
\hline\hline
\multicolumn{2}{|c|}{\textbf{Avg.\ $\Delta$R@1 (RGB$-$Gray)}} &
\multicolumn{6}{c|}{-5.9} &
\multicolumn{6}{c|}{-4.7} &
\multicolumn{6}{c|}{-3.7} \\
\hline
\end{tabularx}
\vspace*{-1em}
\end{table*}
\begin{table}[!ht]
\notsotiny
\centering
\caption{Recall@1 across datasets with MixVPR trained \& evaluated per modality, \textbf{Gray ImageNet pretrained}.}
\label{tab:rest_GrayP}
\setlength{\firstcolw}{1.3cm}
\setlength{\tabcolsep}{2.9pt}
\renewcommand{\arraystretch}{1.0}
\begin{tabular}{|p{\firstcolw}|c|cccc|c|}
\hline
\textbf{Model} &
\textbf{Mode} &
\makecell{\textbf{Pitts250k} \\ \textbf{(Test)}} &
\textbf{SPED} &
\textbf{GLDv2} &
\textbf{Tokyo247} &
\makecell{\textbf{Avg.$\Delta$R@1} \\ \textbf{(RGB-Gray)}} \\
\hline\hline
\multirow{2}{*}{\makecell[l]{\textbf{MixVPR} \\ \textbf{(ResNet18)}}}
& RGB  & \textbf{94.1} & 82.7 & \textbf{50.3} & 74.9  & \multirow{2}{*}{-0.1} \\
& Gray & 93.5 & \textbf{83.7} & 48.8 & \textbf{76.2} & \\
\hline
\multirow{2}{*}{\makecell[l]{\textbf{MixVPR} \\ \textbf{(ResNet34)}}}
& RGB  & 94.8 & \textbf{84.2} & \textbf{51.5} & 76.2  & \multirow{2}{*}{-0.3} \\
& Gray & \textbf{95.0} & 84.0 & \textbf{51.5} & \textbf{77.5} & \\
\hline
\multirow{2}{*}{\makecell[l]{\textbf{MixVPR} \\ \textbf{(ResNet50)}}}
& RGB  & 95.4 & 85.8 & 57.5 & \textbf{86.7}  & \multirow{2}{*}{-0.6} \\
& Gray & \textbf{95.5} & \textbf{87.6} & \textbf{57.9} & \textbf{86.7} & \\
\hline
\end{tabular}
\vspace{-1em}
\end{table}

\begin{enumerate}
    \item \textbf{Nordland:} A $729km$ railway traverse in Norway recorded across spring, summer, fall and winter, evaluating performance under seasonal change. We regularly sample each season traverse to 3,975 frames and remove stationary and tunnel segments.
    \item \textbf{Oxford RobotCar:} Car traverses under six chosen conditions (sun, dusk, rain, overcast, night, snow). We establish 4,054 frame-to-frame correspondences per set at approximately $1m$ spatial sub-sampling.
    \item \textbf{Pitts250k-Test:} 3,498 urban city locations, each with 12 images at 30° intervals. 1000 random queries are sampled, assessing viewpoint and illumination robustness.
    \item \textbf{SPED-Test:} Consists of mixed outdoor scenes captured from fixed cameras under significant illumination, weather, season and viewpoint changes. We use 608 reference-query pairs with one-to-one correspondences.
    \item \textbf{Google Landmarks Dataset (GLDv2):} A landmark-focused benchmark with 23,294 reference and 3,103 queries exhibiting viewpoint, scale and appearance variation, reflecting uncontrolled acquisition.
    \item \textbf{Tokyo24/7:} 6,334 reference locations in Tokyo, each with 12 images at 30° intervals, and 105 query triplets, each containing day, evening and night variants under dense structural change.
\end{enumerate}

\textbf{Retrieval.} Performance is measured with Recall@1, defined by the proportion of queries where the top retrieved image falls within the ground truth tolerance: $5m$ for Oxford RobotCar, $25m$ for Pitts250k and Tokyo24/7, one-to-one correspondence for Nordland and SPED, and provided labels for GLDv2. All training uses an Nvidia A100 GPU with fixed seeds.

\section{ANALYSIS OF RESULTS}
\label{sec:results}
\setlength{\tablew}{\textwidth}
\setlength{\firstcolw}{1.8cm}
{
\begin{table*}[!ht]
\footnotesize
\centering
\caption{Recall@1 across Nordland season pairs with MixVPR trained \& evaluated on the same modality, \textbf{without pretraining}.}
\label{tab:nordland_scratch}
\setlength{\tabcolsep}{4.0pt}
\renewcommand{\arraystretch}{1.0}
\begin{tabular}{|p{\firstcolw}|c|cccc|cccc|cccc|}
\hline
& \textbf{Model} $\rightarrow$ &
\multicolumn{4}{c|}{\textbf{MixVPR -- ResNet18}} &
\multicolumn{4}{c|}{\textbf{MixVPR -- ResNet34}} &
\multicolumn{4}{c|}{\textbf{MixVPR -- ResNet50}} \\
\cline{2-14}
\makecell[l]{\textbf{Reference} $\downarrow$} &
\makecell[b]{\textbf{Query} $\rightarrow$ \\[0.35em] \textbf{Mode} $\downarrow$} &
\rot{\textbf{Fall}} & \rot{\textbf{Spring}} & \rot{\textbf{Summer}} & \rot{\textbf{Winter}} &
\rot{\textbf{Fall}} & \rot{\textbf{Spring}} & \rot{\textbf{Summer}} & \rot{\textbf{Winter}} &
\rot{\textbf{Fall}} & \rot{\textbf{Spring}} & \rot{\textbf{Summer}} & \rot{\textbf{Winter}} \\
\hline\hline
\multirow{2}{*}{\textbf{Fall}}
& RGB  & --- & 88.3 & 92.3 & 62.1 & --- & 86.1 & 92.5 & 59.7 & --- & 83.2 & 90.8 & 52.9 \\
& Gray & --- & \textbf{89.3} & \textbf{94.2} & \textbf{64.9} & --- & \textbf{87.0} & \textbf{93.0} & \textbf{61.8} & --- & \textbf{86.2} & \textbf{92.5} & \textbf{56.5} \\
\hline

\multirow{2}{*}{\textbf{Spring}}
& RGB  & 78.3 & --- & 67.5 & 67.3 & 75.8 & --- & 66.9 & 66.0 & 73.7 & --- & 64.9 & 56.1 \\
& Gray & \textbf{82.2} & --- & \textbf{75.7} & \textbf{78.7} & \textbf{79.2} & --- & \textbf{73.1} & \textbf{75.8} & \textbf{76.7} & --- & \textbf{68.4} & \textbf{70.9} \\
\hline

\multirow{2}{*}{\textbf{Summer}}
& RGB  & 93.5 & 81.4 & --- & 58.9 & \textbf{93.1} & 80.5 & --- & 57.3 & \textbf{92.6} & 79.0 & --- & 50.9 \\
& Gray & \textbf{94.0} & \textbf{83.9} & --- & \textbf{64.6} & 92.4 & \textbf{81.6} & --- & \textbf{62.0} & 92.5 & \textbf{79.8} & --- & \textbf{54.9} \\
\hline

\multirow{2}{*}{\textbf{Winter}}
& RGB  & 49.0 & 64.4 & 44.1 & --- & 52.9 & 68.3 & 49.2 & --- & 39.4 & 53.3 & 36.2 & --- \\
& Gray & \textbf{60.8} & \textbf{75.9} & \textbf{58.2} & --- & \textbf{56.2} & \textbf{74.7} & \textbf{56.5} & --- & \textbf{47.3} & \textbf{68.2} & \textbf{45.9} & --- \\
\hline\hline
\multicolumn{2}{|c|}{\textbf{Avg.\ $\Delta$R@1 (RGB$-$Gray)}} &
\multicolumn{4}{c|}{-6.3} &
\multicolumn{4}{c|}{-3.8} &
\multicolumn{4}{c|}{-5.6\textbf{}} \\
\hline
\end{tabular}
\vspace*{-1em}
\end{table*}
}
\setlength{\tablew}{\textwidth}
\setlength{\firstcolw}{1.5cm}
\begin{table*}[!ht]
\footnotesize
\centering
\caption{Recall@1 across Oxford RobotCar weather pairs with MixVPR trained \& evaluated on the same modality, \textbf{without pretraining}.}
\label{tab:ORC_scratch}
\setlength{\tabcolsep}{1.9pt}
\renewcommand{\arraystretch}{1.0}
\begin{tabularx}{\tablew}{|p{\firstcolw}|c|YYYYYY|YYYYYY|YYYYYY|}
\hline
& \textbf{Model} $\rightarrow$ &
\multicolumn{6}{c|}{\textbf{MixVPR -- ResNet18}} &
\multicolumn{6}{c|}{\textbf{MixVPR -- ResNet34}} &
\multicolumn{6}{c|}{\textbf{MixVPR -- ResNet50}} \\
\cline{2-20}
\makecell[l]{\textbf{Reference} $\downarrow$} &
\makecell[b]{\textbf{Query} $\rightarrow$ \\[0.35em] \textbf{Mode} $\downarrow$} &
\rot{\textbf{Sun}} & \rot{\textbf{Dusk}} & \rot{\textbf{Overcast}} & \rot{\textbf{Rain}} & \rot{\textbf{Night}} & \rot{\textbf{Snow}} &
\rot{\textbf{Sun}} & \rot{\textbf{Dusk}} & \rot{\textbf{Overcast}} & \rot{\textbf{Rain}} & \rot{\textbf{Night}} & \rot{\textbf{Snow}} &
\rot{\textbf{Sun}} & \rot{\textbf{Dusk}} & \rot{\textbf{Overcast}} & \rot{\textbf{Rain}} & \rot{\textbf{Night}} & \rot{\textbf{Snow}} \\
\hline\hline
\multirow{2}{*}{\textbf{Sun}}
& RGB  & --- & 52.0 & 94.4 & 92.5 & 27.1 & 81.7 & --- & 49.3 & 95.1 & 90.9 & 21.4 & 82.2 & --- & 45.0 & 95.8 & \textbf{93.7} & 22.1 & 84.2 \\
& Gray & --- & \textbf{56.7} & \textbf{95.1} & \textbf{93.5} & \textbf{49.6} & \textbf{84.6} & --- & \textbf{55.5} & \textbf{95.4} & \textbf{92.7} & \textbf{48.8} & \textbf{84.5} & --- & \textbf{57.7} & \textbf{96.0} & 93.6 & \textbf{55.4} & \textbf{85.1} \\
\hline

\multirow{2}{*}{\textbf{Dusk}}
& RGB  & 48.1 & --- & 48.9 & 45.0 & 21.4 & 51.2 & 47.6 & --- & 46.6 & 44.6 & 17.6 & 50.1 & 41.1 & --- & 37.5 & 39.1 & 13.2 & 44.9 \\
& Gray & \textbf{52.3} & --- & \textbf{52.1} & \textbf{48.6} & \textbf{39.0} & \textbf{55.7} & \textbf{52.6} & --- & \textbf{52.7} & \textbf{48.3} & \textbf{38.0} & \textbf{53.9} & \textbf{50.8} & --- & \textbf{52.0} & \textbf{48.3} & \textbf{38.6} & \textbf{54.4} \\
\hline

\multirow{2}{*}{\textbf{Overcast}}
& RGB  & \textbf{95.2} & 53.8 & --- & 92.2 & 29.5 & 84.1 & 94.7 & 50.4 & --- & 91.0 & 25.6 & 84.4 & 95.2 & 47.6 & --- & 93.4 & 25.1 & 86.2 \\
& Gray & 94.5 & \textbf{58.5} & --- & \textbf{93.8} & \textbf{56.3} & \textbf{87.8} & \textbf{95.2} & \textbf{58.2} & --- & \textbf{93.4} & \textbf{54.0} & \textbf{87.8} & \textbf{95.8} & \textbf{59.1} & --- & \textbf{94.3} & \textbf{58.8} & \textbf{87.7} \\
\hline

\multirow{2}{*}{\textbf{Rain}}
& RGB  & 88.3 & 46.7 & 89.6 & --- & 22.9 & 73.8 & 87.5 & 44.8 & 90.0 & --- & 22.1 & 73.7 & 90.6 & 43.2 & 91.2 & --- & 18.8 & 76.3 \\
& Gray & \textbf{92.7} & \textbf{49.6} & \textbf{92.6} & --- & \textbf{46.0} & \textbf{79.3} & \textbf{93.0} & \textbf{49.9} & \textbf{92.5} & --- & \textbf{44.9} & \textbf{80.2} & \textbf{93.1} & \textbf{50.6} & \textbf{93.6} & --- & \textbf{48.2} & \textbf{81.2} \\
\hline

\multirow{2}{*}{\textbf{Night}}
& RGB  & 35.0 & 31.5 & 35.3 & 29.7 & --- & 27.8 & 30.6 & 31.0 & 32.7 & 31.9 & --- & 28.0 & 34.0 & 27.5 & 34.0 & 28.8 & --- & 28.1 \\
& Gray & \textbf{57.6} & \textbf{47.6} & \textbf{61.7} & \textbf{52.3} & --- & \textbf{52.7} & \textbf{55.8} & \textbf{46.5} & \textbf{59.8} & \textbf{52.2} & --- & \textbf{48.3} & \textbf{57.8} & \textbf{45.7} & \textbf{60.5} & \textbf{54.7} & --- & \textbf{53.6} \\
\hline

\multirow{2}{*}{\textbf{Snow}}
& RGB  & 84.1 & 60.3 & 88.2 & 79.3 & 26.3 & --- & 83.5 & 56.7 & 87.2 & 78.2 & 24.0 & --- & 87.6 & 55.5 & 90.7 & 82.0 & 21.4 & --- \\
& Gray & \textbf{85.8} & \textbf{61.9} & \textbf{89.8} & \textbf{83.4} & \textbf{45.2} & --- & \textbf{87.5} & \textbf{62.1} & \textbf{89.8} & \textbf{82.5} & \textbf{45.8} & --- & \textbf{87.7} & \textbf{64.8} & \textbf{91.0} & \textbf{84.7} & \textbf{53.6} & --- \\
\hline\hline
\multicolumn{2}{|c|}{\textbf{Avg.\ $\Delta$R@1 (RGB$-$Gray)}} &
\multicolumn{6}{c|}{-9.3} &
\multicolumn{6}{c|}{-10.3} &
\multicolumn{6}{c|}{-12.5} \\
\hline
\end{tabularx}
\vspace*{-1em}
\end{table*}
\begin{table}[!ht]
\notsotiny
\centering
\caption{Recall@1 across datasets with MixVPR trained \& evaluated on the same modality, \textbf{without pretraining}.}
\label{tab:rest_scratch}
\setlength{\firstcolw}{1.3cm}
\setlength{\tabcolsep}{2.9pt}
\renewcommand{\arraystretch}{1.0}
\begin{tabular}{|p{\firstcolw}|c|cccc|c|}
\hline
\textbf{Model} &
\textbf{Mode} &
\makecell{\textbf{Pitts250k} \\ \textbf{(Test)}} &
\textbf{SPED} &
\textbf{GLDv2} &
\textbf{Tokyo247} &
\makecell{\textbf{Avg.$\Delta$R@1} \\ \textbf{(RGB-Gray)}} \\
\hline\hline
\multirow{2}{*}{\makecell[l]{\textbf{MixVPR} \\ \textbf{(ResNet18)}}}
& RGB  & \textbf{91.5} & 72.7 & \textbf{40.2} & \textbf{62.2}  & \multirow{2}{*}{0.2} \\
& Gray & 91.4 & \textbf{78.3} & 39.2 & 56.8 & \\
\hline
\multirow{2}{*}{\makecell[l]{\textbf{MixVPR} \\ \textbf{(ResNet34)}}}
& RGB  & \textbf{90.8} & 72.0 & \textbf{39.5} & \textbf{56.2}  & \multirow{2}{*}{0.4} \\
& Gray & 89.9 & \textbf{76.8} & 37.8 & 52.4 & \\
\hline
\multirow{2}{*}{\makecell[l]{\textbf{MixVPR} \\ \textbf{(ResNet50)}}}
& RGB  & 90.3 & 66.9 & \textbf{41.3} & \textbf{54.6}  & \multirow{2}{*}{-1.4} \\
& Gray & \textbf{90.8} & \textbf{73.1} & 40.0 & \textbf{54.6} & \\
\hline
\end{tabular}
\vspace{-2em}
\end{table}
\textbf{1) Off-the-shelf Performance.} \Cref{tab:nordland_OTS,tab:ORC_OTS,tab:rest_OTS} report the performance of off-the-shelf models on RGB and grayscale versions of each benchmark. We find that the more recent transformer-based models exhibit a significantly smaller performance gap between RGB and gray retrieval, with EigenPlaces achieving an average 7.6\% Recall@1 improvement on Nordland using RGB, while BoQ, on average, only differs by 0.3\%. This is interesting, given the models have been entirely trained on RGB imagery, yet the apparent utility of color to VPR itself is limited. We suspect that EigenPlaces and MixVPR develop a greater reliance on chromatic features under ImageNet training and are unable to adapt across color space, while the DINOv2-based models are sufficiently appearance-invariant through their self-supervised pre-training that color proves to be more redundant for VPR. However, certain conditions in weather or datasets appear to consistently benefit under RGB across the models, such as GLDv2 by $\approx$3\%, suggesting that color may indeed provide some discriminative value when stable chromatic cues are present (GLDv2 is landmark-focused under broadly consistent lighting). Nevertheless, the overall trend suggests that luminance-based structure largely dominates stable global place recognition.

\begin{figure*}[!ht]
  \centering
  \includegraphics[width=0.9\textwidth]{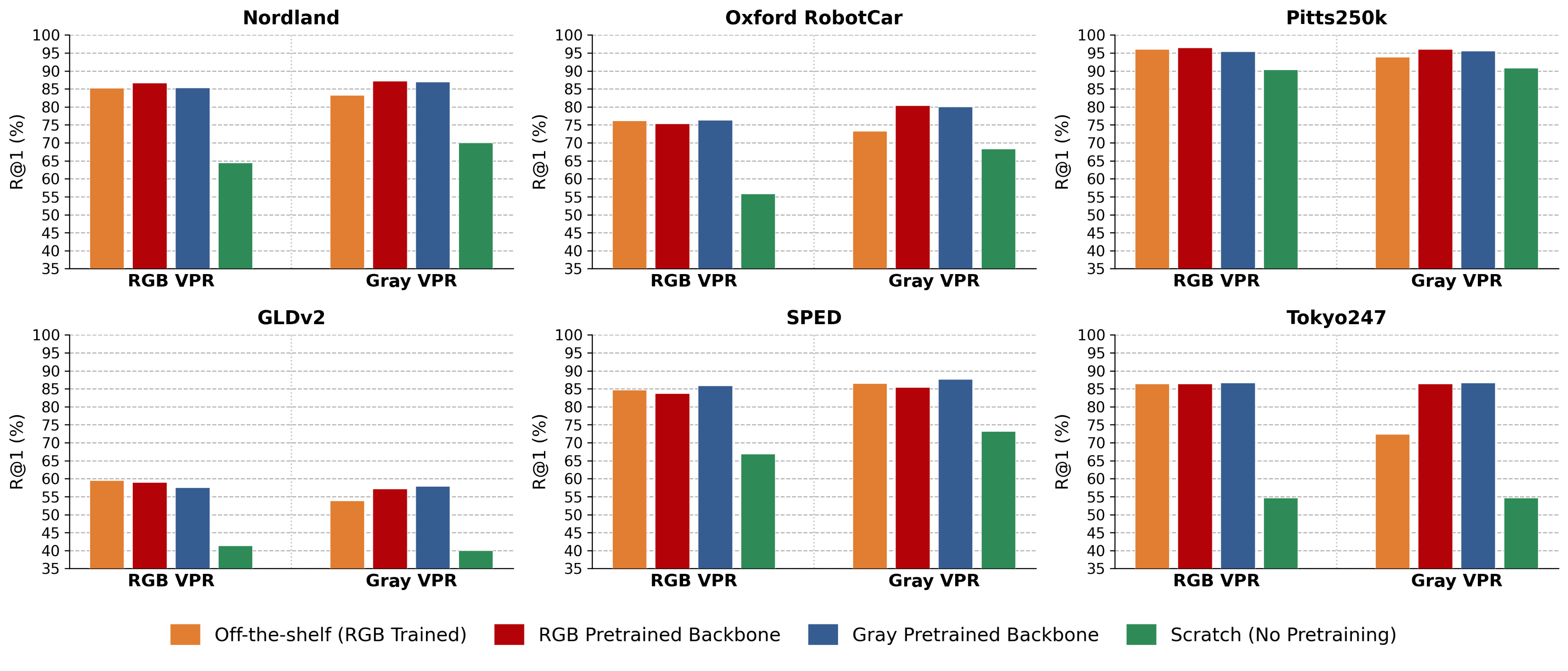}
  \vspace*{-1em}
  \caption{Recall@1 on MixVPR (ResNet50) across the four setups from \fig{fig:blockdiag}. Nordland and Oxford RobotCar results are averaged across season and condition pairs respectively. RGB/Gray VPR denote the training and evaluation modality, except setup 1 where off-the-shelf models are evaluated per modality only.}
  \label{fig:mixvprcomp}
  \vspace*{-1em}
\end{figure*}

\textbf{2) RGB-Pretrained MixVPR.} When controlling for VPR fine-tuning (see \Cref{tab:nordland_RGBP,tab:ORC_RGBP,tab:rest_RGBP}), the performance gap between RGB and grayscale VPR narrows considerably and often reverses in favor of grayscale. While off-the-shelf MixVPR shows more reliance on color, this advantage largely disappears with grayscale VPR fine-tuning, with the model able to perform well under both color spaces and even outperforming RGB under more severe appearance shifts. For example, MixVPR-ResNet50 achieves 82.7\% R@1 on Oxford RobotCar (night reference, overcast query) under grayscale, compared to 71.4\% for the RGB equivalent, as well as outperforming the off-the-shelf model, indicating that its RGB training fails to achieve full color invariance under challenging VPR conditions. Notably, MixVPR-ResNet18 fine-tuned on grayscale sometimes outperforms its ResNet50 RGB counterpart, demonstrating that the robustness gains from grayscale can compensate for a shallower model. Pitts250k and Tokyo24/7 show little sensitivity to VPR input modality, likely reflecting their focus on viewpoint variation over appearance shift, where color remains more consistent. In contrast, SPED consistently benefits from grayscale VPR, while GLDv2 consistently favors RGB, again highlighting that chromatic utility is dependent on the stability and prevalence of color cues in the environment. Generally, from \fig{fig:mixvprcomp}, we see that RGB training offers no significant overall advantage over grayscale across MixVPR configurations.

\textbf{3) Gray-Pretrained MixVPR.} With a grayscale-pretrained backbone, grayscale VPR fine-tuning consistently outperforms its RGB counterpart across all datasets and backbone sizes (\Cref{tab:nordland_GrayP,tab:ORC_GrayP,tab:rest_GrayP}), with the most pronounced gains on appearance-challenging conditions such as Nordland winter and Oxford RobotCar night. As in setup 2, we see that grayscale VPR with ResNet18 can outperform RGB VPR with ResNet50. From \fig{fig:mixvprcomp}, fully grayscale-trained MixVPR performs competitively and can outperform both off-the-shelf and fully RGB-trained variants across benchmarks. However, with GLDv2, RGB-pretraining maintains a marginal advantage of up to 2\%, consistent with its stable landmark-oriented color cues. These results demonstrate that the advantage of color is context-dependent, but limited, making fully grayscale-trained models a compelling and robust alternative for VPR.

\textbf{4) MixVPR - No Pretraining} As expected, removing ImageNet pretraining reduces performance across all cases (see \fig{fig:mixvprcomp}), emphasizing the value of low-level features learned through foundation model pretraining. Nevertheless, from  \Cref{tab:nordland_scratch,tab:ORC_scratch,tab:rest_scratch}, grayscale VPR-only training often outperforms its RGB counterpart, particularly on appearance-challenging datasets, averaging up to 6.3\% R@1 gain on Nordland and 12.5\% on Oxford RobotCar. This suggests that GSV-Cities alone, despite its scale, does not provide sufficient color invariance for robust generalization under the severe appearance shifts present in VPR benchmarks. By removing color, the model is encouraged to prioritize luminance-based structure during training, which remain more stable across appearance conditions. Following the previous setups, performance on Pitts250k remains largely unaffected by input modality, while SPED, GLDv2 and Tokyo24/7 show more variable responses, reflecting greater sensitivity to input modality without the stability provided from pretraining, and a trade-off between structural cues and consistent chromatic features present within the data.

\section{DISCUSSION}
\label{sec:disc}

\textbf{Structure is sufficient for global VPR.} Across evaluations, grayscale models match or outperform their RGB counterparts, with even fully RGB-trained ViT models differing minimally under color removal, reinforcing that luminance-based structure is the driving factor in global place recognition.

\textbf{Color volatility can mislead models.} Color is inherently unstable under appearance shifts, and RGB training can encourage misleading correlations between chromatic cues and places (models without as-extensive pretraining, such as MixVPR, become over-reliant on color features), underperforming their more robust grayscale versions under extreme conditions. Stronger pretraining like in BoQ stabilizes performance across both modes, with grayscale remaining competitive throughout.

\textbf{Color is context-dependent.} Color benefit, generally within 1-4\% R@1, is observed in landmark-focused data where color cues are consistent and prevalent with fewer appearance changes. Nevertheless, structure remains the dominant factor of robust performance.

\textbf{Computational Implications.} Color imagery incurs three times the storage and data movement cost of grayscale. This is increasingly relevant as recent VPR method rely on large-scale training. Datasets such as GSV-Cities ($\sim$23GB), MSLS (196GB cache requirement for NetVLAD representations \cite{Ali-Bey_2022_BMVC}) and SF-XL ($\sim$1TB) highlight the overhead that increases when color invariance is incomplete and augmentation or dataset ensembling is required.

During training, GPU memory is dominated by activations and gradients. Large models such as MegaLoc (228M parameters) report 60GB VRAM usage, requiring batch processing with continuous host-to-device transfer. Such operations make data loading a known bottleneck. While forward and backward pass costs are largely unaffected by channel reduction, single-channel data reduces storage and transfer volume. Across multiple MixVPR training runs, we measured a consistent 29\% reduction in peak RAM utilization across all ResNet backbone sizes and $\sim$1.5$\times$ lower CPU utilization, reflecting reduced loader overhead from decoding, decompressing and transferring batches, and leaving headroom for larger batch sizes or concurrent processes.

Grayscale is also particularly well-suited to TinyML and resource-constrained deployment \cite{10284551}, where micro-controllers operate under 256KB-1MB SRAM budgets. Reducing input tensor size lowers buffering and data movement costs, consistent with benchmarks such as Visual Wake Words and FashionMNIST which use low-resolution or single-channel inputs. Our results show that smaller grayscale-trained backbones can match or outperform larger RGB variants, enabling a practical model-size trade-off. Grayscale further aligns with low-bit embedded sensor modalities common in robotics, such as monochrome, thermal, depth and event cameras, with monochrome cameras additionally avoiding demosaicing overhead and offering improved light sensitivity under low-light conditions that challenge VPR systems.

\section{CONCLUSION}
\label{sec:conclusion}
We investigate the role of color in VPR across training regimes and model architectures. RGB imagery offers limited advantage over grayscale under current SotA conventions, with structure and texture forming the foundation of global VPR performance. Grayscale can even outperform RGB under severe appearance shifts where color is volatile and invariance insufficiently learned. Where persistent chromatic cues exist, RGB may provide support, which is more relevant for landmark-focused retrieval. For deployment, grayscale reduces storage and transfer overhead, improving training efficiency and suiting resource-constrained platforms, making it a practical and compelling default modality for VPR.

\textbf{Limitations and Future Work.} Extending this analysis to transformer-based training remains important, though models such as BoQ already show near-identical RGB and grayscale performance off-the-shelf, suggesting a grayscale-trained variant could likely match of surpass RGB under challenging conditions. Future work includes real-world deployment evaluations across hardware settings and further investigation into model and training design under grayscale input.






\bibliographystyle{IEEEtran} 
\bibliography{IEEEabrv,references}

\end{document}